\documentclass[letterpaper]{article} 
\usepackage{aaai2026}
\usepackage{multirow}
\usepackage{times}  
\usepackage{helvet}  
\usepackage{courier}  
\usepackage[hyphens]{url}  
\usepackage{graphicx} 
\usepackage{natbib}  
\usepackage{caption} 
\usepackage{amsfonts}
\usepackage{amsmath}
\usepackage{algorithm}
\usepackage{algpseudocode}
\usepackage{comment}
\usepackage{makecell}
\usepackage{float}

\usepackage{enumitem}
\usepackage{newfloat}
\usepackage{listings}
\DeclareCaptionStyle{ruled}{labelfont=normalfont,labelsep=colon,strut=off} 
\floatstyle{ruled}
\newfloat{listing}{tb}{lst}{}
\floatname{listing}{Listing}
\title{Discovery of Feasible 3D Printing Configurations for Metal Alloys \\ via AI-driven Adaptive Experimental Design}

\author {
    Azza Fadhel\textsuperscript{\rm 1},
    Nathaniel W. Zuckschwerdt\textsuperscript{\rm 2},
    Aryan Deshwal\textsuperscript{\rm 3},
    Susmita Bose\textsuperscript{\rm 2}, \\
    Amit Bandyopadhyay\textsuperscript{\rm 2},
    Jana Doppa\textsuperscript{\rm 1}
}
\affiliations {
    \textsuperscript{\rm 1}School of Electrical Engineering and Computer Science, Washington State University,\\
    \textsuperscript{\rm 2}School of Mechanical and Materials Engineering, Washington State University,\\
    \textsuperscript{\rm 3}Department of Computer Science and Engineering, University of Minnesota, Twin Cities\\
    \{azza.fadhel, n.zuckschwerdt, sbose, amitband, jana.doppa\}@wsu.edu,\ 
    adeshwal@umn.edu

}

\begin{document}

\maketitle

\begin{abstract}
Configuring the parameters of additive manufacturing processes for metal alloys is a challenging problem due to complex relationships between input parameters (e.g., laser power, scan speed) and quality of printed outputs. The standard trial-and-error approach to find feasible parameter configurations is highly inefficient because validating each configuration is expensive in terms of resources (physical and human labor) and the configuration space is very large. This paper combines the general principles of AI-driven adaptive experimental design with domain knowledge to address the challenging problem of discovering feasible configurations. The key idea is to build a surrogate model from past experiments to intelligently select a small batch of input configurations for validation in each iteration. To demonstrate the effectiveness of this methodology, we deploy it for Directed Energy Deposition process to print GRCop-42, a high-performance copper–chromium–niobium alloy developed by NASA for aerospace applications. Within three months, our approach yielded multiple defect-free outputs across a range of laser powers—dramatically reducing time-to-result and resource expenditure compared to several months of manual experimentation by domain scientists with no success. By enabling high-quality GRCop-42 fabrication on readily available infrared laser platforms for the first time, we democratize access to this critical alloy, paving the way for cost-effective, decentralized production for aerospace applications.

\end{abstract}

\begin{figure*}[t]
    \centering
    \includegraphics[width=0.55\textwidth]{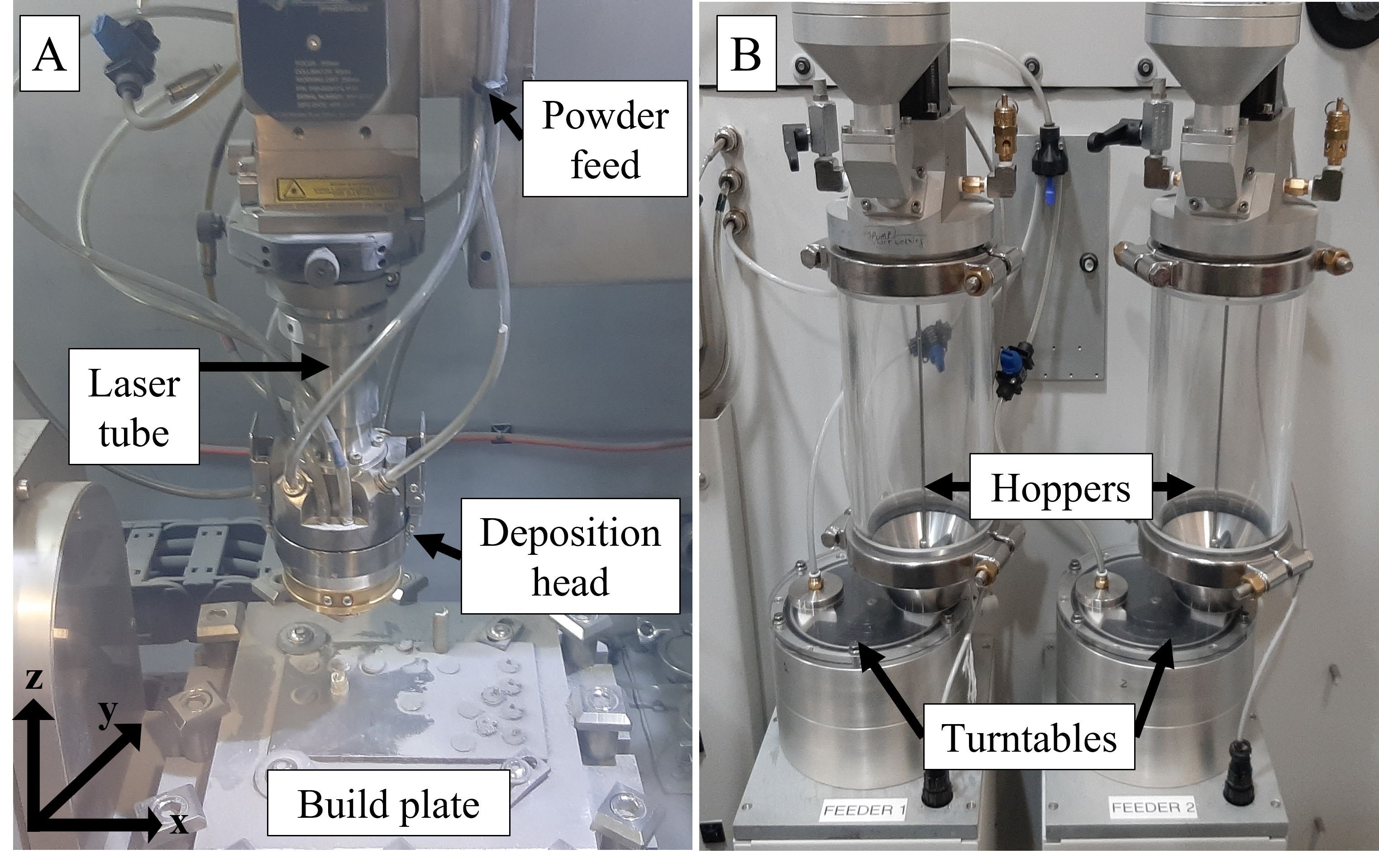}
    \caption{Illustration of FormAlloy Directed energy deposition (DED)-based metal additive manufacturing (AM) system used for our deployment: 
    (A) Build chamber showing the deposition head, powder feeding, and build plate; 
    (B) Two powder feeders.}
    \label{fig:DED}
    \vspace{-3ex}
\end{figure*}

\section{Introduction}

Additive manufacturing (AM) also known as 3D printing, is an important technology that offers cost-effective, sustainable, and highly customizable solutions. There are three key benefits of AM. First, the ability to create highly complex and customized parts by minimizing material waste and production cost \cite{bandyopadhyay2018invited, jung2023additive, landi2022comparative}. AM builds components layer by layer based on digital models, using only necessary material. Second, AM provides flexibility and speed as design changes can be made quickly to enable fast prototyping and product development cycles \cite{jung2023additive, briard2020g}. Third, AM reduces inventory needs and supply chain complexity because it enables on-demand manufacturing and localized production \cite{jung2023additive, briard2020g, doubrovski2011optimal}. 

Despite these major benefits, AM for metal alloys (a combination of two or more metallic elements) presents significant challenges \cite{tofail2018additive, bandyopadhyay2022alloy,bandyopadhyay2022additive}. Due to the complex \underline{unknown} relationships between material, process parameters, and quality of printed outputs, it takes a significant amount of time and resources to identify {\em feasible process parameters}, i.e., input parameters that allow us to successfully print high-quality outputs \cite{bandyopadhyay2018invited, guo2019microstructure}. Directed Energy Deposition (DED) is a popular AM process for metal alloys where process parameter configuration is extremely challenging \cite{svetlizky2021directed}. Achieving consistent microstructure and mechanical properties across the build is a significant metallurgical challenge due to rapid solidification and repeated thermal cycling \cite{bandyopadhyay2018invited, chepiga2023process}. For example, using improper values for parameters such as laser power, scanning speed, and material feed rate can lead to excessive heat input, resulting in low-quality outputs. Similarly, increasing laser power may require adjustments in scanning speed and feed rate to maintain a stable melt pool and prevent defects \cite{bandyopadhyay2018invited, liu2025preparation, chepiga2023process}. 

The standard {\em trial-and-error} approach to uncover feasible process parameters for a given metal alloy is extremely slow which takes several months, and is extremely expensive in terms of both physical resources (cost of materials and equipment for printing) and human labor (to conduct quality analysis of printed outputs for microstructure and mechanical properties) due to the large space of parameter configurations, e.g, $\approx$ over 100 million in DED process \cite{bandyopadhyay2018invited, liu2025preparation,tofail2018additive,bandyopadhyay2022alloy}. The main research, development, and deployment question of this paper is: {\em Can we use AI methods to discover feasible process parameters for a given metal alloy in a resource-efficient manner?}

This paper develops a {\em {\bf B}ayesian {\bf E}xperimental design for {\bf AM} (aka BEAM) approach} and demonstrates its effectiveness to print GRCop-42, a high-performance
copper–chromium–niobium alloy developed by NASA for extreme-temperature aerospace applications \cite{bandyopadhyay2018invited, clare2022alloy}. BEAM is inspired by work on active search \cite{jiang2018efficient} and integrates domain knowledge for improved effectiveness. The key idea is to build a probabilistic surrogate model from past experiments --- pairs of process parameter configuration (input) and binary output indicating the success/failure of the experiment including printing and quality analysis (output) --- and use it to intelligently select a batch of process parameter configurations to try in each iteration. Unlike the trial-and-error approach, this is an {\em active approach within a closed feedback loop} that iterates between the following steps: (a) conducting a batch of experiments using selected parameter configurations , (b) updating our belief about the parameter-feasibility relationship, and (c) selecting the parameter configurations for the next batch of experiments. 

We deployed BEAM in a real-world laboratory with the DED-based AM process to print GRCop-42 \cite{bandyopadhyay2018invited}. 
GRCop-42 offers a unique combination of properties, including high temperature strength, good oxidation resistance, and exceptional resistance to thermal fatigue, making it particularly suitable for use in high-heat flux applications such as rocket engine components \cite{clare2022alloy}. {\em Whether GRCop-42 can be successfully printed at lower laser power levels below 1000W or not is a big open problem}.
Within three months, our BEAM approach yielded multiple defect-free outputs across a range of laser powers—significantly  reducing time-to-result and resource expenditure compared to several months of manual experimentation by our collaborators with no success \cite{chepiga2023process}.  By enabling high-quality GRCop-42 fabrication on readily available infrared laser platforms for the \underline{first-time}, we democratize access to this critical alloy, paving the way for cost-effective, decentralized production of rocket engine chambers, heat exchangers, and other high-heat-flux components \cite{bandyopadhyay2018invited, sheikh2024exploring, tofail2018additive}. Our BEAM approach can be employed to configure AM processes for other metal alloys, and the insights from our deployment can be used to further improve the effectiveness of BEAM for future use-cases \cite{chiappetta2024data, chiappetta2023sparse, deneault2025preferential}.

\vspace{1ex}

\noindent {\bf Contributions.} The key contribution of this paper is the development, deployment, and evaluation of the BEAM approach to efficiently configure the parameters of AM processes for metal alloys. Specific contributions include:
\begin{itemize}
    \item Formulating the discovery of feasible parameter configurations as an adaptive experimental design problem.
    \item Development of the BEAM approach inspired by active search work and incorporating domain knowledge to select a batch of parameter configurations in each iteration.
    \item Deployment and evaluation of BEAM to configure the DED process to print GRCop-42 at lower laser power on  readily available infrared laser platforms for the \underline{first-time} to enable democratization to the masses.
\end{itemize}

\section{Background and Challenges}

In this section, we provide the background on the Directed Energy Deposition process, general challenges for printing metal alloys, and specific challenges for printing GRCop-42, which is the focus of our deployment efforts.

\subsection{AM Process: Directed Energy Deposition}

Directed Energy Deposition (DED) is a typical AM process for metal alloys that uses focused thermal energy, typically from a high-powered laser, to fuse materials by melting them as they are deposited \cite{bandyopadhyay2018additive,debroy2018additive}. DED involves feeding metal in the form of powder or wire directly into the path of the energy source. As the material is simultaneously fed and melted, it bonds layer by layer to form components of the part being built. The DED process is typically carried out using multi-axis machines, which allow for complex geometries.

Fig. \ref{fig:DED} illustrates the DED system employed in our deployment and experimental evaluation. Our system, a 5-axis powder-fed laser deposition setup, operates with a 1000W fiber laser and dual powder hoppers for co-deposition. The process is conducted in an inert argon environment with oxygen levels typically below 20~ppm. The system provides control over four critical processing parameters: {\em laser power, scan speed, powder feed rate, and carrier gas flow rate}. 
The powders used—Inconel 718 and GRCop-42—were purchased from Carpenter Additive \cite{feltner2025particle}, with particle sizes in the 15–53~µm range. Build plates were made from 316L stainless steel (100~mm~$\times$~150~mm~$\times$~2.5~mm), sourced from OnlineMetals.com.

\subsection{General Challenges of Printing Metal Alloys}

Bimetallic structures, which exploit the distinct material properties of two alloys, have significant applications in aerospace \cite{padture2002thermal, onuike2018additive}, automotive \cite{karamics2012effects}, and biomedical industries \cite{balla2010direct}. Conventionally, such structures are fabricated using multi-step joining techniques such as electron beam welding, laser welding \cite{chen2011fibre}, brazing \cite{feng2012reliability}, friction stir welding \cite{fazel2010joining}, and diffusion bonding \cite{elrefaey2009solid}. These methods suffer from high processing costs, long lead times, large heat-affected zones, and defects such as porosity and weak interfaces \cite{bandyopadhyay2018additive,debroy2018additive}. DED-based AM offers a more flexible and rapid route to fabricating such structures \cite{debroy2018additive,khairallah2020controlling}, but brings its own set of challenges.

Specifically, the trial-and-error approach for identifying suitable AM process parameters is both time-consuming and cost-intensive. For example, Inconel 718 costs approximately \$200/kg, while GRCop-42 costs approximately \$300/kg \cite{gradl2019grcop}. Each run on the DED system (including operation, setup, and post-processing) ranges from \$500–\$1000, and post-print quality analysis—including scanning electron microscopy, X-ray diffraction, and mechanical testing—can take 1–4 weeks and add another \$500 per sample \cite{onuike2018additive}. Thus, the overall cost of experimental exploration of the process parameter space can easily exceed \$100,000 and can take significant time, ranging from several months for process optimization of novel alloys.

\subsection{Specific Challenges of Printing GRCop-42}

GRCop-42 is a high-performance copper alloy originally developed by NASA for regeneratively cooled rocket engine combustion chambers \cite{gradl2019grcop}. Its key advantages include high thermal conductivity, creep resistance, and adequate strength at elevated temperatures \cite{ellis2000thermophysical}. However, it poses serious challenges for laser-based metal AM processes such as DED:

\begin{itemize}
    \item \textbf{Low power absorptivity:} Copper reflects most of the infrared spectrum used in typical fiber lasers, making energy absorption inefficient \cite{auwal2018review}.
    \item \textbf{High thermal conductivity:} The rapid heat dissipation prevents localized melting---critical for stable layer-by-layer deposition \cite{freudenberger2018copper}.
    \item \textbf{Uncharted process parameters space:} There is no well-established process window for GRCop-42 using sub-900W lasers in a DED setup \cite{gradl2019grcop}.
    \item \textbf{High cost of trial-and-error:} Each process parameter configuration tested in the lab incurs high material, time, and labor costs, which quickly become unsustainable \cite{onuike2018additive}.
\end{itemize}

The specific goal for our deployment and experimental evaluation was to determine whether GRCop-42 could be printed successfully at laser powers below 900W, ideally in the 500–700W range, using our 1000W laser-DED system. This would significantly reduce energy requirements and open the door to printing with more accessible, lower-power AM systems, i.e., democratization of printing GRCop-42.

\section{Problem Formulation}

We are given an additive manufacturing system (e.g., DED) with $d$ process parameters (e.g., laser power, scan speed, powder feed rate, carrier gas flow rate for DED). Our goal is to discover feasible process parameter configurations to successfully print a given metal alloy (e.g., GRCop-42) by conducting expensive real-world experiments in a physical lab. Each experiment corresponds to evaluating the feasibility of a candidate process parameter configuration (i.e., specific values to parameters) and consists of printing which involves physical resources such as materials/equipment and human labor to perform quality analysis of the printed output to evaluate its microstructure/mechanical properties. Brute-force (trying all possible parameter configurations) and standard trial-and-error approaches are expensive, inefficient, and/or impractical due to two reasons: large space of parameter configurations (e.g., $\approx$ over 100 Million in DED) and experimenting with each candidate configuration is expensive in terms of the cost of resources. Our goal is to accelerate the discovery of feasible parameter configurations by using AI to guide the selection of the sequence of parameter configurations for experimental validation. The specific deployment goal is to identify a diverse set of feasible process parameters that enable successful printing of GRCop-42 using a DED system at low laser power levels (500–700W). 


Formally, Let $\mathcal{X} \subset \mathbb{R}^d$ denote the $d-$dimensional parameter space of controllable AM process variables, and let  $f: \mathcal{X} \rightarrow \{0,1\}$ represent the unknown feasibility function, where $f(\mathbf{x})$ = 1 indicates a successful print configuration and $f(\mathbf{x})$ = 0 represents failure. Given a limited experimental budget $T$, our objective is to discover as many feasible process parameter configurations as possible:
\begin{align}
\max_{\mathcal{S} \subset \mathcal{X}, |\mathcal{S}| \leq T} \sum_{\mathbf{x} \in \mathcal{S}} f(\mathbf{x})
\end{align}

Our formulation shifts the problem away from optimization (i.e., finding maxima/minima) toward one of experimental design and feasibility discovery in a large, sparsely populated space. Specifically, only a small fraction of the parameter space results in successful prints, making the problem more analogous to discovering valid regions in a highly imbalanced setting (number of feasible configurations $\ll$ number of failed configurations).




\section{Related Work}

Bayesian Optimization (BO) is an instantiation of adaptive experimental design that has emerged as a dominant framework for optimizing expensive black-box functions \cite{shahriari2016taking,snoek2012practical,deshwal2023bayesian,BOPS,deshwal2021combining,MerCBO,HyBO,belakaria2019max,belakaria2020uncertainty} and has seen successes in diverse applications including hyperparameter tuning, robotics, and materials design \cite{lookman2019active,deshwal2021bayesian,gantzler2023multi}. BO employs a probabilistic surrogate to model the true objective function and an acquisition strategy to guide evaluation towards promising regions of the input space for the optimization goal \cite{frazier2018tutorial}.

In the context of additive manufacturing, BO has been successfully applied to optimize design parameters that maximize part quality, mechanical strength, or minimize porosity \cite{zhang2020bayesian}. These methods focus on optimizing performance metrics by efficiently exploring the design space with as few experiments as possible. 

However, our problem fundamentally differs from this optimization setting. First, our goal is not to find an optimal configuration, but rather to identify a broad set of feasible configurations to successfully print using a novel alloy (GRCop-42). This shifts the formulation from optimization to discovery, i.e., identifying feasible regions within a high-dimensional and largely infeasible parameter space \cite{toscano2018bayesian,zhao2021active}. Second, BO assumes that the objective function is continuous and that good solutions are clustered near optima. In contrast, the feasible parameter space in our problem is highly sparse and discontinuous, with very few ``successful'' outcomes amid a vast number of failed configurations. 

Our use-inspired challenge is more aligned with a classification problem under severe class imbalance, where successful prints (positive class) are rare and randomly scattered across the input space. This formulation demands alternative strategies such as active learning for imbalanced data and exploration-aware discovery methods \cite{zhao2021active,settles2009active}, rather than global optimization. Prior work on active search \cite{jiang2018efficient} directly addresses our problem formulation. We integrate domain knowledge into active search to improve its effectiveness for our AM application.


\section{AI-Guided Experimental Design to Discover Feasible Parameters}



In this section, we describe the {\em {\bf B}ayesian {\bf E}xperimental design for {\bf AM} (aka BEAM) approach} by building on the general principle of adaptive experimental design. We first provide a high-level overview of BEAM and then explain the key elements that drive the adaptive experimental strategy.

\subsection{Overview of BEAM Approach}

\begin{figure}[h!]
    \centering
    \includegraphics[width=0.45\textwidth]{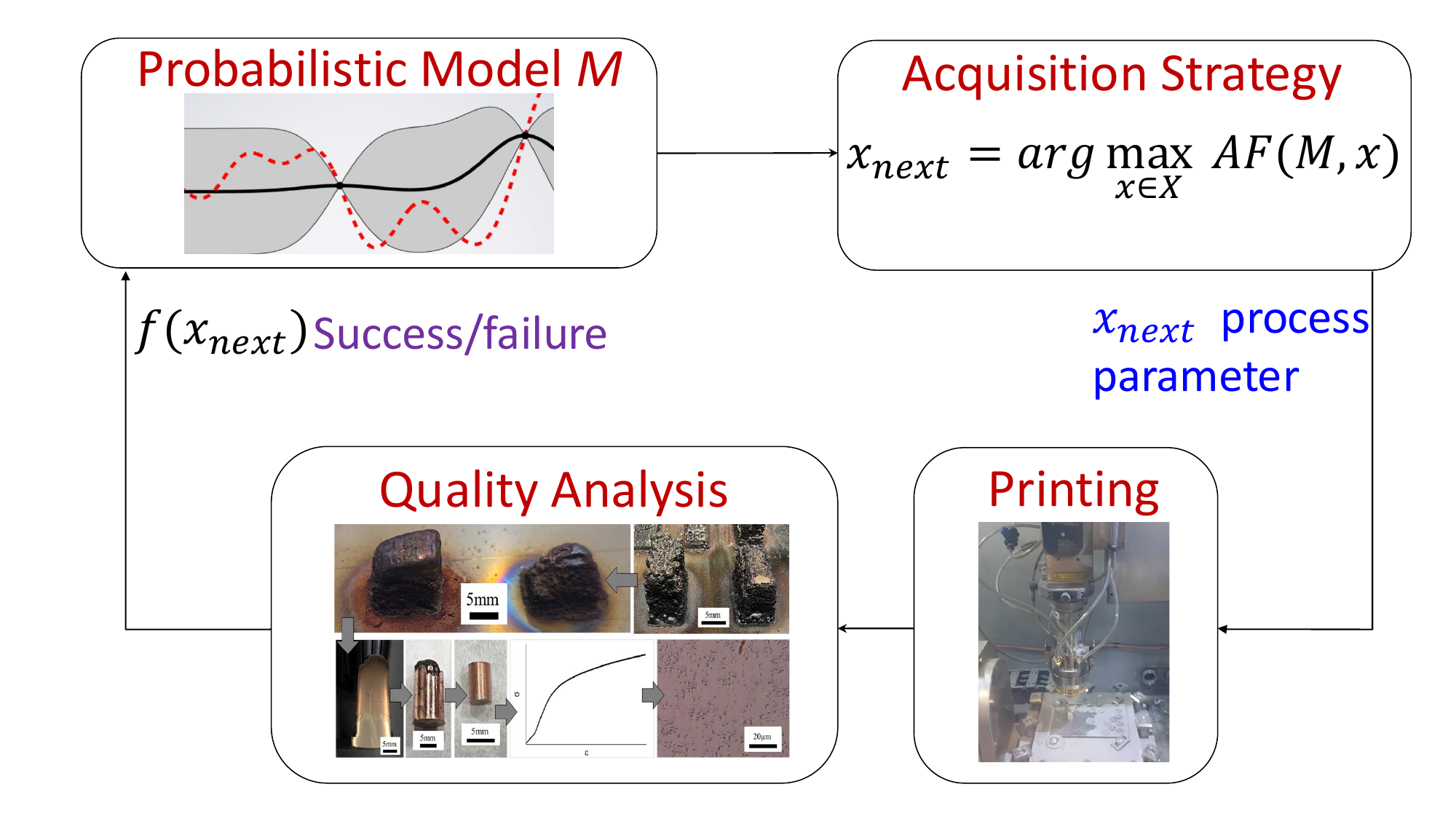} 
    \caption{High-level overview of AI-guided BEAM approach to select one parameter configuration for experimental validation in each iteration (batch size $B$=1). Experiment corresponds to printing using the selected configuration followed by quality analysis to determine its success or failure.}
    \label{fig:your-label}
\end{figure}
BEAM approach for discovering feasible configurations relies on the principles behind adaptive experimental design and is inspired by active search \cite{jiang2018efficient}. Unlike the trial-and-error approach, this is an {\em active approach within a closed feedback loop} that iterates between the following steps: (a) conducting a batch of experiments using selected parameter configurations, (b) updating our belief about the parameter-feasibility relationship, and (c) selecting the parameter configurations for the next batch of experiments. The key idea is to build a probabilistic surrogate model from past experiments --- pairs of process parameter configuration (input) and binary output indicating the success/failure of the experiment including printing and quality analysis (output) --- and use it to intelligently select a batch of process parameter configurations to try in each iteration.

\begin{algorithm}[b!]
\caption{AI-Guided Discovery Framework (BEAM)}
\begin{algorithmic}[1]
\State \textbf{Input:} Process parameter space $\mathcal{X}$, experimental budget $T$, batch size $B$
\State \textbf{Initialize:} $\mathcal{D}_0 \leftarrow$ Initial set of diverse parameter configurations with observed success/failure outcomes
\For{$t = 1$ to $T$}
    \State Train surrogate model on $\mathcal{D}_{t-1}$
    \State Estimate $P(f(\mathbf{x}) = 1)$ for all $\mathbf{x} \in \mathcal{X} \setminus \mathcal{D}_{t-1}$
    \State Select batch of configurations $\{\mathbf{x}_t^{(1)}, \ldots, \mathbf{x}_t^{(B)}\}$ using acquisition strategy to maximize the overall utility
    \ForAll{$\mathbf{x} \in \{\mathbf{x}_t^{(1)}, \ldots, \mathbf{x}_t^{(B)}\}$}
        \State Run experiment and observe $f(\mathbf{x}) \in \{0, 1\}$
        \State Aggregate dataset: $\mathcal{D}_t \leftarrow \mathcal{D}_{t-1} \cup \{(\mathbf{x}, f(\mathbf{x}))\}$
    \EndFor
\EndFor
\State \textbf{Output:} Feasible configurations $\{ \mathbf{x} \in \mathcal{D}_T \mid f(\mathbf{x}) = 1 \}$
\end{algorithmic}
\label{algo:BEAM}
\end{algorithm}

The overall closed feedback loop of experimental workflow for BEAM consists of the following key steps.

\begin{enumerate}
    \item \textbf{Initialization:} A small number of diverse parameter configurations validated using experiments. This can be done using Latin hypercube sampling when there is no prior knowledge or can be selected using expert intuition and physical constraints. In our deployment, we employed the data from all the failed experiments by our AM collaborators, which was the starting point for this project.
    
    \item \textbf{Surrogate modeling:} A probabilistic machine learning model is trained on the labeled data collected so far, to capture our current belief about the relationship between process parameters and feasibility. Importantly, this surrogate model can provide both its prediction and uncertainty for unknown (i.e., not evaluated yet) configurations which are used to select configurations for validation.
    
    \item \textbf{Intelligent selection of a batch of configurations:} The acquisition strategy identifies the most promising parameter configurations for the next batch of $B \geq 1$ experiments, considering uncertainty, coverage, and feasibility.
    
    \item \textbf{Experimental validation:} The selected configurations are validated by conducting experiments with a real AM system, and the observed outcomes (success/failure) are incorporated into the aggregated training set.
\end{enumerate}

This closed feedback loop repeats until the experimental budget $T$ is exhausted. By using AI to guide experimental trials than trial-and-error or brute-force search, our BEAM approach significantly improves the resource-efficiency to discover feasible parameter configurations. Algorithm \ref{algo:BEAM} provides a high-level pseudo-code of BEAM.

\subsection{Key Elements of BEAM}

To efficiently explore the process parameter space and reduce the number of failed experiments using BEAM, we need to instantiate the two key elements: 1) Surrogate model to capture the current belief about the relationship between parameters and feasibility; and 2) Acquistion strategy that selects which new batch of parameter configurations to evaluate in each iteration guided by the surrogate model. 
We describe the details of these two elements below: 

\vspace{1ex}

{\noindent \bf Surrogate Model:} At iteration $t$, having observed data from prior printing experiments, $\mathcal{D}_t$ = $\{(\mathbf{x}_i, y_i)\}_{i=1}^t$ where $y_i$ = $f(\mathbf{x}_i)$, we employ a probabilistic classifier as a surrogate model for the underlying feasibility function. Our surrogate model provides posterior distribution for a candidate $x$ (i.e., probability of success for any candidate configuration) which intelligently guides the decision making procedure:
\begin{align}
p_t(\mathbf{x}) = P(f(\mathbf{x}) = 1 | \mathcal{D}_t)
\end{align}

We utilize a probabilistic variant of k-nearest neighbors \cite{jiang2018efficient, garnett2012bayesian} as our surrogate model.

\vspace{1ex}

{\noindent \bf Acquisition Strategy:} To select the next experiments under a limited budget, we use a model-guided decision policy from active search \cite{jiang2018efficient} which addresses the key challenge that feasible configurations are sparsely distributed across the parameter space. A naive approach to candidate selection would simply optimize the immediate posterior probability $p_t(\mathbf{x})$, selecting:
\begin{equation}
\mathbf{x}_{t+1} = \arg\max_{\mathbf{x} \in \mathcal{X} \setminus \mathcal{D}_t} p_t(\mathbf{x})
\end{equation}

However, this greedy strategy is fundamentally flawed because it fails to reason about the available experimental budget and focuses purely on exploitation. Such myopic behavior can lead to premature convergence around known feasible regions while leaving large portions of the parameter space unexplored. A more principled approach must balance exploitation of current model knowledge with exploration that improves understanding of unseen regions. Therefore, we employ an acquisition function that explicitly combines these objectives:
\begin{equation}
\alpha_t(\mathbf{x}) = p_t(\mathbf{x}) + \beta_t(\mathbf{x})
\end{equation}
where $p_t(\mathbf{x})$ represents the \textbf{exploitation term} and $\beta_t(\mathbf{x})$ captures the \textbf{exploration term}.

The exploration term $\beta_t(\mathbf{x})$ quantifies the expected value of improved model knowledge for the remaining experimental budget \cite{jiang2018efficient}. Specifically, it estimates how evaluating candidate $\mathbf{x}$ would enhance our ability to identify feasible configurations in subsequent experiments. We approximate this exploration value by considering how evaluating $\mathbf{x}$ would affect the sum of feasibility probabilities across the remaining budget. After observing the outcome $y$ at location $\mathbf{x}$, the model's posterior beliefs will be updated, potentially changing feasibility estimates for all unexplored candidates. The exploration term captures this expected improvement:
\begin{equation}
\beta_t(\mathbf{x}) = \mathbb{E}_{y \sim \text{Bernoulli}(p_t(\mathbf{x}))} \left[ \sum_{j=1}^{T-t-1} \max_{\mathbf{x}' \in \mathcal{X} \setminus \mathcal{D}_{t+1}} p_{t+1}^{(j)}(\mathbf{x}') \right]
\end{equation}
where $p_{t+1}^{(j)}(\mathbf{x}')$ denotes the $j$-th highest feasibility probability among unexplored candidate configurations under the updated posterior $P(f(\cdot) = 1 \mid \mathcal{D}_t \cup \{(\mathbf{x}, y)\})$.

Direct computation of this expectation is computationally intractable. Therefore, we approximate it by assuming that the remaining budget $(T-t-1)$ would be optimally allocated to the configurations in a single batch \cite{jiang2018efficient}:
\begin{equation}
\beta_t(\mathbf{x}) \approx \mathbb{E}_{y} \left[ \sum_{j=1}^{T-t-1} p_{t+1}^{(j)} \right]
\end{equation}

\begin{align}
\mathbf{x}_{t+1} = \arg\max_{\mathbf{x} \in \mathcal{X} \setminus \mathcal{D}_t} \alpha_t(\mathbf{x}) \label{eqn:afo}
\end{align}

\vspace{1ex}

{\bf \noindent Integration of Domain Knowledge:} Solving the search problem in Equation (\ref{eqn:afo}) is expensive. In order to improve the computational/sample efficiency of BEAM, we embed expert knowledge through hard constraints on the parameter ranges and physical feasibility checks, which helps in {\em sound pruning} of the search space and prevents from wasting experimental resources on infeasible configurations.


\begin{table*}[h!]
\centering
{\footnotesize
\renewcommand{\arraystretch}{1.3} 
\begin{tabular}{|c|c|c|c|c|}
\hline
\makecell{Laser Power} & \makecell{Experimental \\budget} & \makecell{Discovery \\ rate} & \makecell{Size of \\ parameter space} & \makecell{Fraction of \\ explored space} \\
\hline
950W & 10 & 1 & $>100$M & $10 \times 10^{-8}$ \\
\hline
700W & 10 & 1 & $>100$M & $10 \times 10^{-8}$ \\
\hline
600W & 10 & 3 & $>100$M & $10 \times 10^{-8}$ \\
\hline
500W & 10 & 1 & $>100$M & $10 \times 10^{-8}$ \\
\hline
\end{tabular}
}
\caption{Results of AI-guided discovery approach (BEAM) for feasible process parameter configurations for different laser power levels. BEAM discovers at least one feasible configuration for each laser power level within a budget of 10 experiments.}
\label{tab:results}
\end{table*}
\section{Experiments and Results}

In this section, we describe our experimental setup for deployment and evaluation of BEAM to discover feasible parameter configurations to print GRCop-42 on a DED system.

\subsection{Experimental Setup and Deployment Details}

\vspace{1ex}

\noindent {\bf AM process and metal alloy.} We deployed and evaluated BEAM through a combination of online AI-guided experiment runs and manual lab validation in a real-world additive manufacturing workflow. We focused on printing the copper-based alloy {\em GRCop-42} on inconel 718 using a commercial {\em directed energy deposition (DED)} system. The goal was to identify feasible parameter configurations that produce successful prints for low laser-power levels (less than 1000W). For each laser power setting (950W, 700W, 600W, and 500W), the search space size is approximately over {100 million unique parameter configurations}. 

\vspace{1ex}

\noindent {\bf Process parameter space.} The search space $\mathfrak{X} \subset \mathbb{R}^{5}$ consists of five controllable DED process parameters. Their domains and discretizations were defined based on machine capabilities and input from our additive manufacturing experts:

\begin{itemize}[itemsep=0.5ex]
    \item \textbf{Feed Rate (RPM):} $[0.01, 1.0]$, interval $0.01$
    \item \textbf{Gas Flow Rate (L/min):} $[3.0, 10.0]$, interval $0.5$
    \item \textbf{Thickness of Inconel:} $[0, 10]$, interval $0.2$
    \item \textbf{Scan Speed (mm/min):} $[200, 1600]$, interval $50$
    \item \textbf{Layer Height (mm):} $[0.05, 0.5]$, interval $0.01$
\end{itemize}

This results in an extremely large combinatorial design space---approximately over 100 million configurations.

To maintain consistency across runs, secondary factors such as part geometry, hatching strategy, and build length were fixed. Hatch spacing was constrained to a maximum of 1 mm. The Interpass Wait was fixed to 0~s.


\vspace{1ex}

\noindent {\bf Implementation details of BEAM.} Recall that BEAM needs a data-driven surrogate model to predict the feasibility (success/failure) of unknown configurations based on past experimental data. We employ a probabilistic variant of $k$-Nearest Neighbors ($k=5$), which estimates the likelihood of success based on local class distributions in the normalized configuration space. We use the training data from 37 failed experiments from our additive manufacturing collaborators to initialize the surrogate model. We selected $B$=2 experiments in each iteration of BEAM until we found a feasible configuration for a fixed laser power level. The trade-off is between using every experimental validation to select the next experiments (resource-efficiency for $B$=1) vs. throughput with larger batches ($B >$ 1).

\subsection{Deployment Goals and Evaluation Metrics}
\label{sec:evaluation-metrics}

\vspace{1ex}

\noindent {\bf Deployment Goals.} GRCop-42 being a copper-based alloy, can be hard to process due to its low laser absorptivity and high thermal conductivity. Thus, when printing GRCop-42, most opt for equipment with a laser power of 2 KW to 4 KW. This greatly limits the number of machines that GRCop-42 can be printed on, as more than 90 percent of the current printers on the market operate with a laser power of 500W to 1000W. The other advantages of printing GRCop-42 at lower laser powers include reduced post-processing costs, lower energy use, extended equipment lifespan, and faster qualification/certification (more details and discussion on these advantages can be found in a later section). However, {\em whether GRCop-42 can be successfully printed at lower laser power levels or not is a big open problem}.

Hence, our goal of deploying AI-guided approach was to discover at least one feasible parameter configuration for each laser power setting (950W, 700W, 600W, 500W) progressively from 950W to 500W to test the limits of feasible printing of GRCop-42 at lower laser power. For each laser power value $\in \{ 950W, 700W, 600W, 500W\}$, we set the experimental budget $T$=10 and increase it by 10 if we are not able to discover feasible configurations within the budget. 

\vspace{1ex}

\noindent {\bf Evaluation Metrics.} We evaluate the effectiveness of AI-guided discovery approach BEAM using two metrics.

\begin{itemize}

\item {\em Experimental resource-efficiency:} measures the efficiency with respect to brute-force search, i.e., performing experiments with all possible parameter configurations. We employ the fraction of experiments/physical resource cost/human labor time needed by BEAM to discover at least one feasible configuration (smaller the better). 

\item {\em Discovery rate:} measures the total number of feasible parameter configurations discovered by BEAM within the given experimental budget (higher the better).

\end{itemize}

\begin{table*}[!h]
\centering
{\footnotesize
\setlength{\tabcolsep}{4pt} 
\renewcommand{\arraystretch}{1.3} 
\begin{tabular}{|c|c|c|c|c|c|c|c|c|c|c|}
\hline
\makecell{Laser Power\\(W)} & 
\makecell{Feed\\(RPM)} & 
\makecell{Gas Flow\\(l/min)} & 
\makecell{Thickness of \\Inconel} & 
\makecell{Scan Speed\\(mm/min)} & 
\makecell{Layer Height\\(mm)} \\
\hline
950 & 0.17 & 7 & 6 & 750 & 0.2 \\
\hline
700 & 0.2 & 7 & 3.8 & 1600 & 0.11 \\
\hline
\multirow{3}{*}{}  & 1 & 10 & 10 & 1600 & 0.49 \\

600 & 0.4 & 7 & 5.4 & 1550 & 0.17\\

& 0.2 & 7 & 7 & 1600 & 0.11 \\
\hline
500 & 0.075 & 7 & 4 & 250 & 0.3 \\
\hline
\end{tabular}
}
\caption{Feasible process parameter configurations discovered by AI-guided BEAM approach for different laser power levels.}
\label{tab:config-values}
\end{table*}

\subsection{Results and Discussion}

\vspace{1ex}

\noindent {\bf AI-guided discovery vs. Brute force search.} Table \ref{tab:results} shows the experimental efficiency and discovery rate of our AI-guided approach BEAM for each of the four laser power levels. Remarkably, BEAM was able to discover at least one feasible parameter configuration for each laser power level we considered within the specified budget of 10 experiments and within a span of three months. BEAM discovered three feasible configurations for laser power level 600W and one feasible configuration each for levels 950W, 700W, and 500W. Table \ref{tab:config-values} shows the values of feasible parameter configurations discovered by BEAM. In contrast, our additive manufacturing collaborators were not able to find even a single feasible configuration after trying 37 parameter configurations across several months. Compared to a brute-force strategy—which would require evaluating over 100 million configurations for each laser power setting, the AI-guided approach BEAM is able to discover at least one feasible configuration within 10 experiments (extremely small fraction of parameter space). This means significant savings in terms of resource cost for materials, human labor cost to conduct quality analysis of printed outputs, and the overall time needed to discover feasible parameter configurations.

The results from BEAM to successfully print GRCop-42 at lower laser power demonstrate the transformative potential of AI-driven adaptive experimental design to accelerate the discovery of feasible parameter configurations for additive manufacturing processes in a resource-efficient manner. 


\vspace{1ex}

\begin{figure}[h!] 
\centering 
\includegraphics[width=0.9\linewidth]{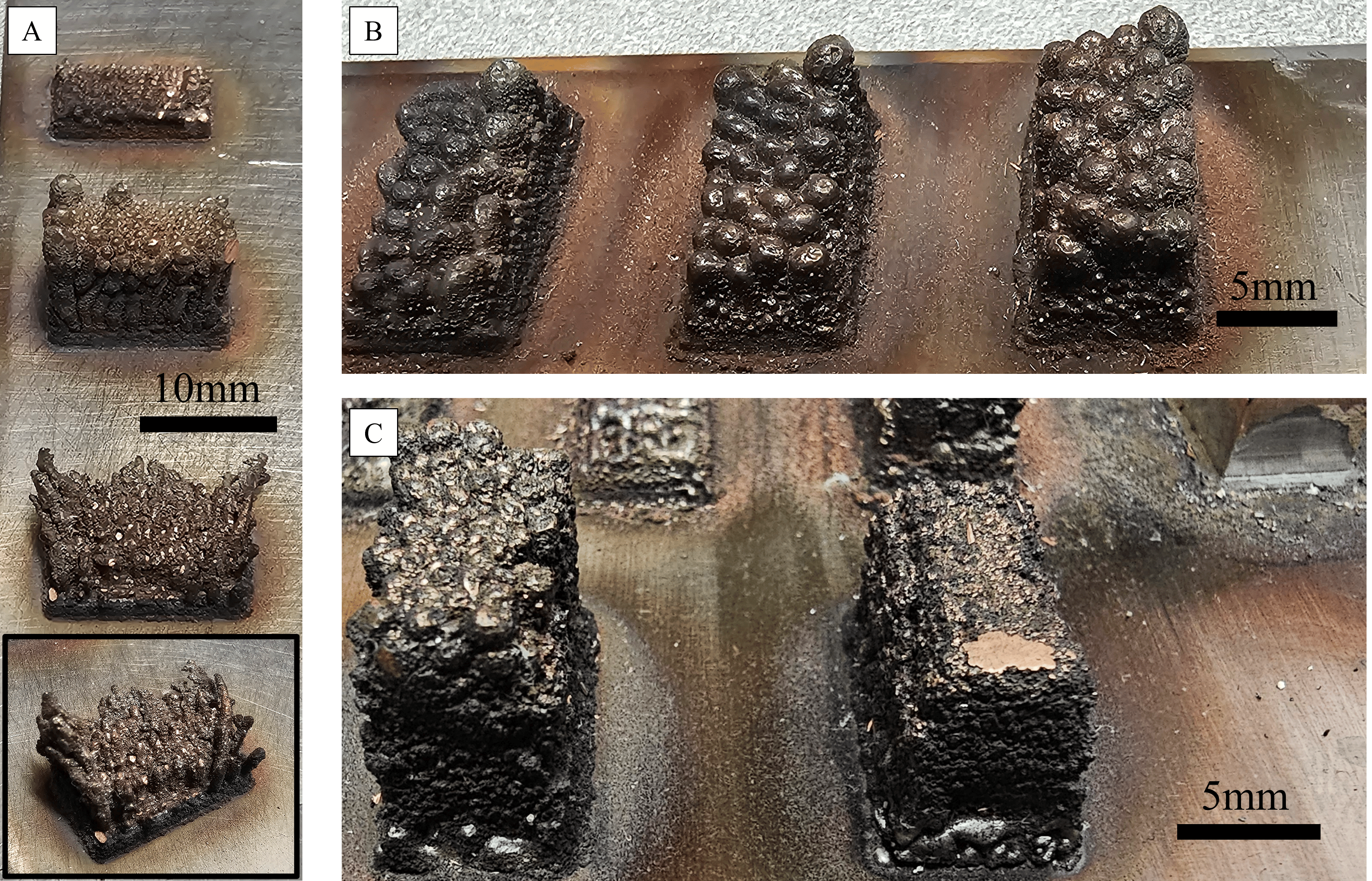} \caption{\textbf{Printed failed structures}; (A) Test blocks with long column structures with the lowest being a tested structure with the columns being broken. (B) Improved test blocks with the column size increasing to much larger pillars. (C) Taller structures with the columns starting higher in the test block.} \label{fig:failed-structures} 
\end{figure}

\noindent {\bf Analysis of failed and successful prints.} Many parameters ended with failed parts, as shown in Fig. \ref{fig:failed-structures}. Prints of the GRCop42 initially were printed like the samples in Figure \ref{fig:failed-structures}(A), with long, thin pillars that easily separated when attempting to machine the samples, causing the samples to fall apart. There were also failures, as shown in Fig. \ref{fig:failed-structures}(B), where the pillars grew in size, resulting in large beads on the top surfaces of the prints. This made it simpler to see if the print would succeed, as these beads would continue to grow upwards, forming the columns at a larger scale. Figure \ref{fig:failed-structures}(C) still failed prints, as attempting to print the samples larger would result in the same structures as seen in Figure \ref{fig:failed-structures}(B).

\begin{figure}[h!] 
\centering 
\includegraphics[width=0.9\linewidth]{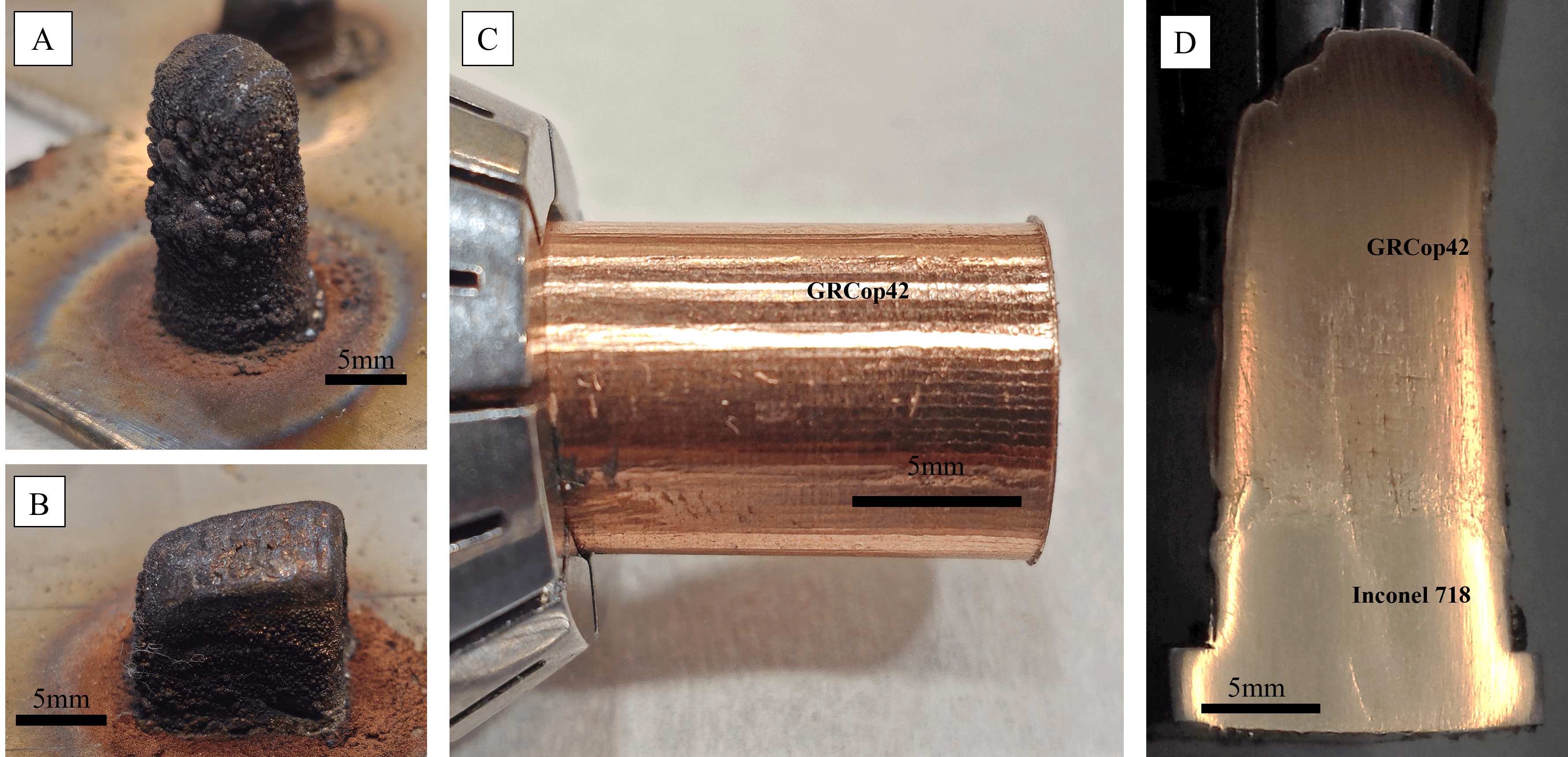} \caption{\textbf{Printed successful structures from 700W laser power using DED-based metal AM}; (A) As-printed cylinder. (B) Printed test block. (C) Machined cylinder with solid structure after removal of exterior shell from printing. (D) Cross-section of as-printed cylinder sample with even transition between the Inconel718 and GRCop42.} \label{fig:700w-ded-structures} 
\end{figure}

Successful print parameters were determined by printing the test blocks first, as seen in Fig. \ref{fig:700w-ded-structures}(B), to determine if a larger surface area shape could be printed. Next, further parameter testing is done by printing cylinders that would be used for mechanical testing, as shown in Fig. \ref{fig:700w-ded-structures}(A). These cylinders were then machined to remove the exterior shell, revealing a solid interior with minimal defects in the material, as shown in Fig. \ref{fig:700w-ded-structures}(C). The transition between the GRCop42 and Inconel 718, seen in Fig. \ref{fig:700w-ded-structures}(D), shows a smooth transition from one material to the next with no defects.


\section{Practical Impact and Deployment Lessons}

This section provides details of the real-world impact of our AI-guided discovery of feasible configurations for AM process and discuss the lessons learned from our deployment.

\subsection{Real-world Impact}

\vspace{1ex}

\noindent {\bf Sustainable jet engines.} GRCop-42 is an alloy that NASA developed for use in jet engines. Current developments are for thin coatings onto the existing jet engine alloys, such as Inconel 718, to allow a higher operating temperature. A temperature increase of 50$^\circ\mathrm{C}$ for the operating temperature of these engines could increase fuel efficiency by up to 10 percent. Thus, our AI-guided discovery of feasible process configurations will help to make planes more efficient in one of the last areas that can be improved, given that planes can be made as light as possible with the structure of the aircraft.

\vspace{1ex}

\noindent {\bf Democratization of efficient GRCop-42 printing.} The feasible process parameter configurations discovered by our AI approach demonstrates for the \underline{first-time} that GRCop-42 can be printed using more than 90 percent of the current printers on the market which operate with laser power levels between 500W to 1000W. With the use of AI, we were able to achieve successful prints within the range of 950W to 500W in 3 months, resulting in significant cost savings in machine time, materials, and consumables. This democratization fosters broader participation across academia, emerging industries, and small research laboratories.

Printing GRCop-42 at \underline{lower laser powers} offers several commercial advantages, especially for industries such as aerospace in terms of reliability and cost-efficiency.

\begin{itemize}
\item {\em Reduced post-processing costs:} Lower laser power can produce parts with better surface finish, reduced distortion, and fewer defects. This minimizes the need for extensive post-processing steps such as machining, heat treatment, or surface polishing—leading to lower overall production costs and faster lead times.

\item {\em Lower energy use and increased lifespan of equipment:} Operating machines at lower laser power can reduce wear on key components (e.g., laser optics) and decrease energy consumption per build. Over time, this can lead to lower maintenance costs and energy savings, making production more sustainable and economically viable.

\item {\em Faster qualification and certification:} Consistent microstructure and minimal variation between builds make it easier to qualify and certify parts for critical applications. This translates to faster time-to-market, especially in regulated industries such as aerospace or defense where certification processes can be time-consuming.

\end{itemize}

\subsection{Lessons Learned from Deployment}

\vspace{1ex}

\noindent {\bf New insights for DED-based manufacturing.} Table \ref{tab:config-values} shows the values of feasible parameter configurations discovered by our AI-guided approach. We uncover two novel domain insights. First, the successful parameter configurations that were determined by the AI algorithm showed that the key was to use a very high scan speed and lower-than-normal layer heights to achieve solid parts. Second, the surprising part was the much higher scan speed that was used for other materials in the DED process, as well as the short amount of time that it took to find usable process parameters. These insights can be useful to try appropriate parameter configurations for other metal alloys and similar AM processes while initializing BEAM for deployment.

\vspace{1ex}

\noindent {\bf Importance of domain knowledge.} The expert knowledge from our additive manufacturing collaborators helped the deployment of AI-guided approach in two ways. First, appropriate discretization of parameter values to reduce the search space without loss in performance/discovery rate. Second, the physical constraints that allowed us to soundly prune some candidate parameter configurations. Whenever such domain knowledge is available, it should be incorporated into the deployment of BEAM for other use-cases.

\vspace{1ex}

\noindent {\bf Consistency of AI-guided approach.} The experimental efficiency to discover feasible parameter configurations can slightly vary from one use-case to another. However, our deployment showed that across four different settings (laser power levels), the AI-guided approach was able to consistently discover feasible configurations within a budget of 10 experiments. This gives reasonable confidence that our observed effectiveness will generalize to other use-cases.

\vspace{1ex}

\noindent {\bf Outlook for GRCop-42 specific results.} Focusing on the impact of our specific results, showing AI can help establish AM-based metallic alloy processability at a lower laser power, it is vital to look at various alloys and the current challenges related to AM processing of them. The most commonly used metallic materials are iron-based steels and aluminum alloys. Although some steels are processed using metallic-AM, most alloy steel compositions are not due to the need for higher laser powers. Similarly, most commercial aluminum alloys, such as Al6061, Al7075, or Al2024, cannot be processed using a regular metal AM system due to laser power limitations. Yet, those alloys are extensively used in the aerospace and automobile industries, making the fuselage of planes to high-end car bodies. An ability to process those alloys within 1000W laser power can unlock the potential for AM operations beyond its current reach and become a more versatile manufacturing platform in the near future. We believe that our work is impactful in making such a dream come true in laser-based metal AM. 

\section{Summary and Future Outlook}

This paper studied the deployment of an AI-guided experimental design approach to discover feasible process parameter configurations for additive manufacturing (AM) of metal alloys. Specifically, we demonstrated the potential of AI-guided discovery to identify parameter configurations of a directed energy deposition system to print metal alloy GRCop-42 at low laser power levels between 1000W and 500W for the first time using a small number of experiments, saving significant resources and time. We outlined the lessons learned from our deployment for printing other metal alloys and/or additive manufacturing processes.

Metals and alloys are used extensively in many demanding applications where high stress or temperatures are common. Innovations in metallic materials are slow, and most metallic parts used today, from aerospace to medical to other industrial applications, use legacy alloys that have been in use for over 100 years. The primary reason for the lack of interest in alloy design comes from the necessity to have extensive processing equipment that can operate at high temperatures (typically, $>$ 2000$^\circ\mathrm{C}
$) and the high cost/time associated with trial-and-error experimentation with no guarantees for success. AI has the potential to accelerate alloy design with the help of metal AM technologies, and that is the primary significance of our work. 
Our work established the processability issues related to AM using AI, and showed that AI is effective when only one composition is used. In the future, many compositions can be tested to replace the legacy alloys with better-performing alloy designs, and we believe that AI will be an integral part of such research. 

\newpage

\section*{Acknowledgements}

Aryan Deshwal is supported by National Science Foundation (NSF) grant IIS-2313174. The views expressed are those of the authors and do not reflect the official policy or position of the NSF.

\bibliography{aaai2026}

@inproceedings{gradl2019grcop,
  title={GRCop-42 development and hot-fire testing using additive manufacturing powder bed fusion for channel-cooled combustion chambers},
  author={Gradl, Paul R and Protz, Christopher S and Cooper, Kenneth and Ellis, David and Evans, Laura J and Garcia, Chance},
  booktitle={AIAA Propulsion and Energy 2019 Forum},
  pages={4228},
  year={2019}
}

@article{snoek2012practical,
  title={Practical bayesian optimization of machine learning algorithms},
  author={Snoek, Jasper and Larochelle, Hugo and Adams, Ryan P},
  journal={Advances in neural information processing systems},
  volume={25},
  year={2012}
}

@article{lookman2019active,
  title={Active learning in materials science with emphasis on adaptive sampling using uncertainties for targeted design},
  author={Lookman, Turab and Balachandran, Prasanna V and Xue, Dezhen and Yuan, Ruihao},
  journal={npj Computational Materials},
  volume={5},
  number={1},
  pages={21},
  year={2019},
  publisher={Nature Publishing Group UK London}
}

@article{bandyopadhyay2018additive,
  title={Additive manufacturing of multi-material structures},
  author={Bandyopadhyay, Amit and Heer, Bryan},
  journal={Materials Science and Engineering: R: Reports},
  volume={129},
  pages={1--16},
  year={2018},
  publisher={Elsevier}
}

@article{padture2002thermal,
  title={Thermal barrier coatings for gas-turbine engine applications},
  author={Padture, Nitin P and Gell, Maurice and Jordan, Eric H},
  journal={Science},
  volume={296},
  number={5566},
  pages={280--284},
  year={2002},
  publisher={American Association for the Advancement of Science}
}

@inproceedings{zhao2021active,
  title={Active learning under label shift},
  author={Zhao, Eric and Liu, Anqi and Anandkumar, Animashree and Yue, Yisong},
  booktitle={International Conference on artificial intelligence and statistics},
  pages={3412--3420},
  year={2021},
  organization={PMLR}
}

@article{onuike2018additive,
  title={Additive manufacturing of Inconel 718--copper alloy bimetallic structure using laser engineered net shaping (LENS)},
  author={Onuike, Benedict and Heer, Bhaskar and Bandyopadhyay, Amit},
  journal={Additive Manufacturing},
  volume={21},
  pages={133--140},
  year={2018},
  publisher={Elsevier}
}

@article{karamics2012effects,
  title={The effects of different ceramics size and volume fraction on wear behavior of Al matrix composites (for automobile cam material)},
  author={Karam{\i}{\c{s}}, Mehmet Baki and Cerit, A Alper and Sel{\c{c}}uk, Burhan and Nair, Fehmi},
  journal={Wear},
  volume={289},
  pages={73--81},
  year={2012},
  publisher={Elsevier}
}

@article{balla2010direct,
  title={Direct laser processing of a tantalum coating on titanium for bone replacement structures},
  author={Balla, Vamsi Krishna and Banerjee, Shashwat and Bose, Susmita and Bandyopadhyay, Amit},
  journal={Acta biomaterialia},
  volume={6},
  number={6},
  pages={2329--2334},
  year={2010},
  publisher={Elsevier}
}

@article{chen2011fibre,
  title={Fibre laser welding of dissimilar alloys of Ti-6Al-4V and Inconel 718 for aerospace applications},
  author={Chen, Hui-Chi and Pinkerton, Andrew J and Li, Lin},
  journal={The International Journal of Advanced Manufacturing Technology},
  volume={52},
  number={9},
  pages={977--987},
  year={2011},
  publisher={Springer}
}

@article{feng2012reliability,
  title={Reliability studies of Cu/Al joints brazed with Zn--Al--Ce filler metals},
  author={Feng, Ji and Songbai, Xue and Wei, Dai},
  journal={Materials \& Design},
  volume={42},
  pages={156--163},
  year={2012},
  publisher={Elsevier}
}

@article{fazel2010joining,
  title={Joining of CP-Ti to 304 stainless steel using friction stir welding technique},
  author={Fazel-Najafabadi, Mahmoud and Kashani-Bozorg, SF and Zarei-Hanzaki, Abbas},
  journal={Materials \& Design},
  volume={31},
  number={10},
  pages={4800--4807},
  year={2010},
  publisher={Elsevier}
}

@article{elrefaey2009solid,
  title={Solid state diffusion bonding of titanium to steel using a copper base alloy as interlayer},
  author={Elrefaey, A and Tillmann, W},
  journal={Journal of materials processing technology},
  volume={209},
  number={5},
  pages={2746--2752},
  year={2009},
  publisher={Elsevier}
}

@article{debroy2018additive,
  title={Additive manufacturing of metallic components--process, structure and properties},
  author={DebRoy, Tarasankar and Wei, Huiliang L and Zuback, James S and Mukherjee, Tuhin and Elmer, John W and Milewski, John O and Beese, Allison Michelle and Wilson-Heid, A de and De, Amitava and Zhang, Wei},
  journal={Progress in materials science},
  volume={92},
  pages={112--224},
  year={2018},
  publisher={Elsevier}
}

@article{khairallah2020controlling,
  title={Controlling interdependent meso-nanosecond dynamics and defect generation in metal 3D printing},
  author={Khairallah, Saad A and Martin, Aiden A and Lee, Jonathan RI and Guss, Gabe and Calta, Nicholas P and Hammons, Joshua A and Nielsen, Michael H and Chaput, Kevin and Schwalbach, Edwin and Shah, Megna N and others},
  journal={Science},
  volume={368},
  number={6491},
  pages={660--665},
  year={2020},
  publisher={American Association for the Advancement of Science}
}

@techreport{ellis2000thermophysical,
  title={Thermophysical properties of GRCop-84},
  author={Ellis, David L and Keller, Dennis J and Nathal, Michael},
  year={2000}
}

@article{auwal2018review,
  title={A review on laser beam welding of copper alloys},
  author={Auwal, ST and Ramesh, Singh and Yusof, Farazila and Manladan, Sunusi Marwana},
  journal={The International Journal of Advanced Manufacturing Technology},
  volume={96},
  number={1},
  pages={475--490},
  year={2018},
  publisher={Springer}
}

@incollection{freudenberger2018copper,
  title={Copper and copper alloys},
  author={Freudenberger, Jens and Warlimont, Hans},
  booktitle={Springer Handbook of Materials Data},
  pages={297--305},
  year={2018},
  publisher={Springer}
}

@article{feltner2025particle,
  title={Particle size and shape analyses for powder bed additive manufacturing},
  author={Feltner, Langdon and Korte, Ethan and Bahr, David F and Mort, Paul},
  journal={Particuology},
  volume={101},
  pages={33--42},
  year={2025},
  publisher={Elsevier}
}

@article{shahriari2016taking,
  title={Taking the human out of the loop: A review of Bayesian optimization},
  author={Shahriari, Bobak and Swersky, Kevin and Wang, Ziyu and Adams, Ryan P and De Freitas, Nando},
  journal={Proceedings of the IEEE},
  volume={104},
  number={1},
  pages={148--175},
  year={2016},
  publisher={IEEE}
}

@article{frazier2018tutorial,
  title={A tutorial on Bayesian optimization},
  author={Frazier, Peter I},
  journal={arXiv preprint arXiv:1807.02811},
  year={2018}
}

@article{zhang2020bayesian,
  title={Bayesian Optimisation for Sequential Experimental Design with Applications in Additive Manufacturing},
  author={Zhang, Parnel and others},
  journal={Additive Manufacturing},
  volume={36},
  pages={101493},
  year={2020},
  publisher={Elsevier}
}

@article{toscano2018bayesian,
  title={Bayesian optimization with expensive integrands},
  author={Toscano-Palmerin, Saul and Frazier, Peter I},
  journal={arXiv preprint arXiv:1803.08661},
  year={2018}
}

@book{settles2009active,
  title={Active Learning Literature Survey},
  author={Settles, Burr},
  year={2009},
  publisher={University of Wisconsin-Madison}
}

@article{jung2023additive,
  title={Is additive manufacturing an environmentally and economically preferred alternative for mass production?},
  author={Jung, Sangjin and Kara, Levent Burak and Nie, Zhenguo and Simpson, Timothy W and Whitefoot, Kate S},
  journal={Environmental science \& technology},
  volume={57},
  number={16},
  pages={6373--6386},
  year={2023},
  publisher={ACS Publications}
}

@article{landi2022comparative,
  title={Comparative life cycle assessment of two different manufacturing technologies: laser additive manufacturing and traditional technique},
  author={Landi, Daniele and Zefinetti, Filippo Colombo and Spreafico, Christian and Regazzoni, Daniele},
  journal={Procedia CIRP},
  volume={105},
  pages={700--705},
  year={2022},
  publisher={Elsevier}
}

@article{briard2020g,
  title={G-DfAM: a methodological proposal of generative design for additive manufacturing in the automotive industry},
  author={Briard, Tristan and Segonds, Fr{\'e}d{\'e}ric and Zamariola, Nicolo},
  journal={International Journal on Interactive Design and Manufacturing (IJIDeM)},
  volume={14},
  number={3},
  pages={875--886},
  year={2020},
  publisher={Springer}
}

@inproceedings{doubrovski2011optimal,
  title={Optimal design for additive manufacturing: opportunities and challenges},
  author={Doubrovski, Zjenja and Verlinden, Jouke C and Geraedts, Jo MP},
  booktitle={International design engineering technical conferences and computers and information in engineering conference},
  volume={54860},
  pages={635--646},
  year={2011}
}

@article{chiappetta2024data,
  title={Data-informed uncertainty quantification for laser-based powder bed fusion additive manufacturing},
  author={Chiappetta, Mihaela and Piazzola, Chiara and Tamellini, Lorenzo and Reali, Alessandro and Auricchio, Ferdinando and Carraturo, Massimo},
  journal={International Journal for Numerical Methods in Engineering},
  volume={125},
  number={17},
  pages={e7542},
  year={2024},
  publisher={Wiley Online Library}
}

@article{chiappetta2023sparse,
  title={Sparse-grids uncertainty quantification of part-scale additive manufacturing processes},
  author={Chiappetta, Mihaela and Piazzola, Chiara and Carraturo, Massimo and Tamellini, Lorenzo and Reali, Alessandro and Auricchio, Ferdinando},
  journal={International Journal of Mechanical Sciences},
  volume={256},
  pages={108476},
  year={2023},
  publisher={Elsevier}
}

@article{chepiga2023process,
  title={Process parameter selection for production of stainless steel 316L using efficient multi-objective Bayesian optimization algorithm},
  author={Chepiga, Timur and Zhilyaev, Petr and Ryabov, Alexander and Simonov, Alexey P and Dubinin, Oleg N and Firsov, Denis G and Kuzminova, Yulia O and Evlashin, Stanislav A},
  journal={Materials},
  volume={16},
  number={3},
  pages={1050},
  year={2023},
  publisher={MDPI}
}

@article{deneault2025preferential,
  title={Preferential Bayesian optimization improves the efficiency of printing objects with subjective qualities},
  author={Deneault, James R and Kim, Woojae and Kim, Jiseob and Gu, Yuzhe and Chang, Jorge and Maruyama, Benji and Myung, Jay I and Pitt, Mark A},
  journal={Digital Discovery},
  volume={4},
  number={3},
  pages={723--737},
  year={2025},
  publisher={Royal Society of Chemistry}
}

@article{bandyopadhyay2018invited,
  title={Invited review article: Metal-additive manufacturing---Modeling strategies for application-optimized designs},
  author={Bandyopadhyay, Amit and Traxel, Kellen D},
  journal={Additive manufacturing},
  volume={22},
  pages={758--774},
  year={2018},
  publisher={Elsevier}
}

@article{clare2022alloy,
  title={Alloy design and adaptation for additive manufacture},
  author={Clare, Adam T and Mishra, Rajiv S and Merklein, Marion and Tan, Hua and Todd, Iain and Chechik, Lova and Li, Jingjing and Bambach, Markus},
  journal={Journal of Materials Processing Technology},
  volume={299},
  pages={117358},
  year={2022},
  publisher={Elsevier}
}

@article{sheikh2024exploring,
  title={Exploring chemistry and additive manufacturing design spaces: a perspective on computationally-guided design of printable alloys},
  author={Sheikh, Sofia and Vela, Brent and Attari, Vahid and Huang, Xueqin and Morcos, Peter and Hanagan, James and Acemi, Cafer and Karaman, Ibrahim and Elwany, Alaa and Arroyave, Raymundo},
  journal={Materials Research Letters},
  volume={12},
  number={4},
  pages={235--263},
  year={2024},
  publisher={Taylor \& Francis}
}

@article{guo2019microstructure,
  title={Microstructure and properties of Cu-Cr-Nb alloy with high strength, high electrical conductivity and good softening resistance performance at elevated temperature},
  author={Guo, Xiaoli and Xiao, Zhu and Qiu, Wenting and Li, Zhou and Zhao, Ziqian and Wang, Xu and Jiang, Yanbin},
  journal={Materials Science and Engineering: A},
  volume={749},
  pages={281--290},
  year={2019},
  publisher={Elsevier}
}

@article{liu2025preparation,
  title={Preparation of GRCop-42 Cu Alloy by Laser-Directed Energy Deposition: Role of Laser Power on Densification, Microstructure, and Mechanical Properties},
  author={Liu, Chao and Han, Ping and Sun, Hongwei and Zhao, Yun},
  journal={Crystals},
  volume={15},
  number={6},
  pages={547},
  year={2025},
  publisher={MDPI}
}

@article{tofail2018additive,
  title={Additive manufacturing: scientific and technological challenges, market uptake and opportunities},
  author={Tofail, Syed AM and Koumoulos, Elias P and Bandyopadhyay, Amit and Bose, Susmita and O'Donoghue, Lisa and Charitidis, Costas},
  journal={Materials today},
  volume={21},
  number={1},
  pages={22--37},
  year={2018},
  publisher={Elsevier}
}

@article{bandyopadhyay2022alloy,
  title={Alloy design via additive manufacturing: Advantages, challenges, applications and perspectives},
  author={Bandyopadhyay, Amit and Traxel, Kellen D and Lang, Melanie and Juhasz, Michael and Eliaz, Noam and Bose, Susmita},
  journal={Materials Today},
  volume={52},
  pages={207--224},
  year={2022},
  publisher={Elsevier}
}

@article{svetlizky2021directed,
  title={Directed energy deposition (DED) additive manufacturing: Physical characteristics, defects, challenges and applications},
  author={Svetlizky, David and Das, Mitun and Zheng, Baolong and Vyatskikh, Alexandra L and Bose, Susmita and Bandyopadhyay, Amit and Schoenung, Julie M and Lavernia, Enrique J and Eliaz, Noam},
  journal={Materials Today},
  volume={49},
  pages={271--295},
  year={2021},
  publisher={Elsevier}
}

@article{bandyopadhyay2022additive,
  title={Additive manufacturing of bimetallic structures},
  author={Bandyopadhyay, Amit and Zhang, Yanning and Onuike, Bonny},
  journal={Virtual and Physical Prototyping},
  volume={17},
  number={2},
  pages={256--294},
  year={2022},
  publisher={Taylor \& Francis}
}

@article{deshwal2021bayesian,
  title={Bayesian optimization of nanoporous materials},
  author={Deshwal, Aryan and Simon, Cory M and Doppa, Janardhan Rao},
  journal={Molecular Systems Design \& Engineering},
  volume={6},
  number={12},
  pages={1066--1086},
  year={2021},
  publisher={Royal Society of Chemistry}
}

@article{gantzler2023multi,
  title={Multi-fidelity Bayesian optimization of covalent organic frameworks for xenon/krypton separations},
  author={Gantzler, Nickolas and Deshwal, Aryan and Doppa, Janardhan Rao and Simon, Cory M},
  journal={Digital Discovery},
  volume={2},
  number={6},
  pages={1937--1956},
  year={2023},
  publisher={Royal Society of Chemistry}
}

@article{belakaria2019max,
  title={Max-value entropy search for multi-objective Bayesian optimization},
  author={Belakaria, Syrine and Deshwal, Aryan and Doppa, Janardhan Rao},
  journal={Advances in neural information processing systems},
  volume={32},
  year={2019}
}

@inproceedings{belakaria2020uncertainty,
  title={Uncertainty-aware search framework for multi-objective Bayesian optimization},
  author={Belakaria, Syrine and Deshwal, Aryan and Jayakodi, Nitthilan Kannappan and Doppa, Janardhan Rao},
  booktitle={Proceedings of the AAAI Conference on Artificial Intelligence},
  volume={34},
  number={06},
  pages={10044--10052},
  year={2020}
}

@article{deshwal2021combining,
  title={Combining latent space and structured kernels for Bayesian optimization over combinatorial spaces},
  author={Deshwal, Aryan and Doppa, Jana},
  journal={Advances in neural information processing systems},
  volume={34},
  pages={8185--8200},
  year={2021}
}

@inproceedings{deshwal2023bayesian,
  title={Bayesian optimization over high-dimensional combinatorial spaces via dictionary-based embeddings},
  author={Deshwal, Aryan and Ament, Sebastian and Balandat, Maximilian and Bakshy, Eytan and Doppa, Janardhan Rao and Eriksson, David},
  booktitle={International Conference on Artificial Intelligence and Statistics},
  pages={7021--7039},
  year={2023},
  organization={PMLR}
}

@inproceedings{HyBO,
  title={Bayesian optimization over hybrid spaces},
  author={Deshwal, Aryan and Belakaria, Syrine and Doppa, Janardhan Rao},
  booktitle={International conference on machine learning},
  pages={2632--2643},
  year={2021},
  organization={PMLR}
}

@inproceedings{MerCBO,
  title={Mercer features for efficient combinatorial Bayesian optimization},
  author={Deshwal, Aryan and Belakaria, Syrine and Doppa, Janardhan Rao},
  booktitle={Proceedings of the AAAI Conference on Artificial Intelligence},
  volume={35},
  number={8},
  pages={7210--7218},
  year={2021}
}

@inproceedings{BOPS,
  title={Bayesian optimization over permutation spaces},
  author={Deshwal, Aryan and Belakaria, Syrine and Doppa, Janardhan Rao and Kim, Dae Hyun},
  booktitle={Proceedings of the AAAI conference on artificial intelligence},
  volume={36},
  number={6},
  pages={6515--6523},
  year={2022}
}

@article{jiang2018efficient,
  title={Efficient nonmyopic active search with applications in drug and materials discovery},
  author={Jiang, Shali and Malkomes, Gustavo and Moseley, Benjamin and Garnett, Roman},
  journal={arXiv preprint arXiv:1811.08871},
  year={2018}
}

@article{garnett2012bayesian,
  title={Bayesian optimal active search and surveying},
  author={Garnett, Roman and Krishnamurthy, Yamuna and Xiong, Xuehan and Schneider, Jeff and Mann, Richard},
  journal={arXiv preprint arXiv:1206.6406},
  year={2012}
}

\end{document}